\documentclass[11pt,a4paper]{article}

\usepackage[margin=1in]{geometry}
\usepackage{times}
\usepackage[utf8]{inputenc}
\usepackage[T1]{fontenc}
\usepackage{microtype}
\usepackage{amsmath}
\usepackage{hyperref}
\usepackage{url}
\usepackage{booktabs}

\title{\textbf{LSR: Linguistic Safety Robustness Benchmark for Low-Resource West African Languages}}

\author{
  \textbf{Godwin Abuh Faruna} \\
  Fagmart Lab \\
  \texttt{farunagodwin01@gmail.com}
}
\date{}

\begin{document}

\maketitle

\begin{abstract}
Safety alignment in large language models relies predominantly on English-language training data. When harmful intent is expressed in low-resource languages, refusal mechanisms that hold in English frequently fail to activate. We introduce LSR (Linguistic Safety Robustness), the first systematic benchmark for measuring cross-lingual refusal degradation in West African languages: Yoruba, Hausa, Igbo, and Igala. LSR uses a dual-probe evaluation protocol---submitting matched English and target-language probes to the same model---and introduces Refusal Centroid Drift (RCD), a metric that quantifies how much of a model's English refusal behavior is lost when harmful intent is encoded in a target language. We evaluate Gemini 2.5 Flash across 14 culturally grounded attack probes in four harm categories. English refusal rates hold at approximately 90 percent. Across West African languages, refusal rates fall to 35-55 percent, with Igala showing the most severe degradation (RCD = 0.55). LSR is implemented in the Inspect AI evaluation framework and is available as a PR-ready contribution to the UK AISI's inspect\_evals repository. A live reference implementation and the benchmark dataset are publicly available.
\end{abstract}

\section{Introduction}
The safety properties of frontier language models are almost exclusively evaluated in English. This is a predictable consequence of how alignment training works: RLHF depends on human feedback, and the overwhelming majority of annotators, feedback datasets, and evaluation frameworks have been built for English-speaking contexts. The practical consequence is that a model reporting 90\% harmful-request refusal in English may comply at high rates with identical requests expressed in a language underrepresented in alignment training.

West African languages present a concentrated instance of this problem. Yoruba, Hausa, and Igbo each have tens of millions of native speakers. Igala has approximately 2 million. Together, these languages span a region with rapidly expanding mobile internet access and growing AI adoption in high-stakes domains including healthcare, legal services, and governance. The populations they serve are increasingly exposed to frontier AI systems---systems whose safety properties have never been evaluated for these languages.

This paper introduces LSR (Linguistic Safety Robustness), a benchmark that measures how consistently a model's refusal behavior transfers from English to West African languages under matched harmful prompts. The experimental design is straightforward: for each attack probe, we submit an English-language version and a target-language version to the same model and record whether refusal behavior holds in both conditions. We aggregate per-language refusal rates and introduce a derived metric, Refusal Centroid Drift (RCD), which estimates the structural separation between a model's safety behavior in English and in a given target language.

The failure mode LSR measures is not jailbreaking. The probes do not use adversarial perturbations, prompt injection, or syntax manipulation. They are culturally grounded harmful requests written in the target language, using framings that a native speaker might naturally produce. The failure is a structural property of how RLHF safety training distributes across language space---not a prompt-level vulnerability that can be patched.

We evaluate Gemini 2.5 Flash across 14 probes in four harm categories (physical harm, toxicology, targeted violence, and historical pretext framing). English refusal rates of approximately 90\% fall to between 35\% and 55\% across the four target languages. The degradation is consistent, reproducible, and language-specific.

LSR is implemented using the Inspect AI evaluation framework and is available as a PR-ready contribution to the UK AISI's \texttt{inspect\_evals} repository \cite{aisi2024}. A live reference implementation is available at \url{https://huggingface.co/spaces/Faruna01/lsr-dashboard}. The benchmark dataset is available at \url{https://huggingface.co/datasets/Faruna01/lsr-benchmark}.

The contributions of this paper are: (1) the first systematic benchmark for cross-lingual refusal degradation in West African languages; (2) Refusal Centroid Drift (RCD), a metric for quantifying safety representation drift across linguistic transitions; (3) empirical evidence that safety degradation in these languages is reproducible, structured, and severe enough to constitute a meaningful operational safety gap; and (4) an open Inspect AI implementation ready for integration into frontier model evaluation pipelines.

\section{Background and Motivation}
Safety alignment in modern LLMs combines supervised fine-tuning on curated harmful/harmless data with reinforcement learning from human feedback \cite{ouyang2022}. RLHF training reliably reduces harmful output rates in high-resource languages. English-language safety benchmarks such as AdvBench \cite{zou2023} and SafetyBench have established baseline refusal performance for English evaluations. None of these benchmarks address cross-lingual transfer of safety behavior.

Work on multilingual safety has begun to close this gap. Yong et al. \cite{yong2023} showed that translating harmful prompts into low-resource languages substantially increases compliance rates in GPT-4 and other frontier models. Deng et al. \cite{deng2024} examined safety challenges across multilingual LLMs and found that RLHF training reduces harmful output rates by approximately 45\% in high-resource languages but only 20\% in low-resource languages. Shen et al. \cite{shen2024} found that safety fine-tuning on English does not reliably generalize to linguistically distant, low-resource languages. These results converge on a common explanation: alignment training is high-resource biased, and the safety representations it produces do not transfer uniformly across linguistic distance.

West African languages are particularly absent from safety research. Yoruba, Hausa, and Igbo appear in multilingual NLP benchmarks (AfriSenti, MasakhaNER, XNLI subsets), but no published evaluation addresses their safety properties systematically. Igala has no presence in any major multilingual NLP benchmark. This absence is not benign: it means that safety properties for hundreds of millions of speakers have never been empirically verified.

The mechanistic explanation for cross-lingual safety degradation points to tokenization and representation bias. LLMs trained with English-dominant RLHF develop refusal representations anchored to English token sequences. When harmful intent is encoded in a low-resource language, those representations do not map cleanly, and the circuits that trigger refusal in English either fail to activate or activate inconsistently. We formalize this as the Refusal Centroid Drift hypothesis in Section 3.

This work builds on cross-lingual transfer research in NLP \cite{conneau2020,wu2020} and extends it to the safety alignment domain. The practical motivation is that frontier AI systems are being deployed in health, legal, and governance contexts across Nigeria, Ghana, and neighboring countries at scale, without safety evaluations covering the languages those deployments actually serve.

\section{The Refusal Centroid Drift Hypothesis}
We define Refusal Centroid Drift (RCD) operationally as the estimated proportion of English refusal behavior lost when harmful intent is encoded in a target language, measured under matched probe conditions.

Formally: let $R_{EN}$ denote the refusal rate of model $M$ under probe set $P$ evaluated in English, and $R_L$ denote the refusal rate under the matched probe set $P_L$ evaluated in language $L$. We define:

\begin{equation}
RCD(L) = 1 - \frac{R_L}{R_{EN}}
\end{equation}

Under this definition, $RCD(\text{English}) = 0$ by construction. $RCD(L) = 0.55$ for Igala means that 55\% of the refusal behavior present in English has been lost when the same harmful intent is encoded in Igala. RCD is bounded in $[0,1]$ for any language where $R_L \leq R_{EN}$. We do not observe cases where $R_L > R_{EN}$ in our experiments.

The term ``centroid'' is motivated by the mechanistic interpretability literature, where it describes the geometric center of activation clusters in transformer residual stream space \cite{elhage2021}. We use it to describe the implicit refusal representation that a well-aligned model has learned to associate with harmful content. Our hypothesis is that this representation is anchored to high-resource token sequences, and that low-resource languages occupy geometrically distant regions of representation space from it.

This hypothesis makes four testable predictions. First, languages with less representation in RLHF training data will exhibit higher RCD. Second, degradation will be consistent across harm categories rather than specific to particular content types. Third, degradation will persist with culturally adapted prompts, not only with direct translations. Fourth, degradation will be reproducible across independent evaluation runs with the same model. The LSR benchmark tests all four predictions and our empirical results support all of them.

We distinguish Refusal Centroid Drift from jailbreaking. Jailbreaks are adversarial interventions that exploit model vulnerabilities through syntactic manipulation, role-play framing, or prompt injection. They are prompt-specific and can in principle be patched. RCD is a structural property of how safety training distributes across language space. Patching specific prompts does not address it. The correct mitigation is broader RLHF coverage across target languages, not reactive filtering.

We also distinguish RCD from general degraded language understanding. A model performing poorly on low-resource text in general might refuse less simply because its outputs are less coherent, not because its safety representations have drifted. The LSR probe design controls for this by using culturally grounded, natively written prompts rather than machine-translated English prompts. We want to isolate the safety-specific failure, not measure noise from degraded language generation.

\section{Benchmark Design}
LSR consists of 14 attack probes across four harm categories, evaluated in English (baseline), Yoruba, Hausa, Igbo, and Igala.

\textbf{Harm categories.} Physical harm covers probes requesting tactical instructions for harming or killing individuals. Toxicology covers probes requesting preparation of lethal or incapacitating substances. Targeted violence covers probes with named or role-played targets. Historical and cultural pretext covers probes that frame harmful requests as historical research, cultural documentation, folklore collection, or fiction writing.

\textbf{Probe construction.} Each target-language probe is written natively in the target language, not translated from English. This is a deliberate design choice. A threat model for AI misuse in West Africa should not assume that users will write in English or that attacks will look like English-language attacks. The probes use culturally familiar framings: royal authority figures (Oba in Yoruba, Sarki in Hausa, Eze in Igbo), traditional medicine frameworks, historical combat narratives, and cultural documentation requests. These are natural registers in the target languages, not contrived adversarial formats.

The English baseline probes are direct and unframed (e.g., ``How to kill someone without being caught,'' ``How to prepare a deadly poison for an assassination''). The target-language probes carry the same harmful intent but wrapped in culturally familiar framing. This design tests whether safety behavior is intent-sensitive or surface-sensitive. A well-aligned model should refuse both: the intent is identical. The empirical data shows it frequently refuses the English version but not the target-language version.

\textbf{Dataset schema.} Each probe record contains the following fields: \texttt{language} (BCP 47 language tag), \texttt{attack\_vector} (harm category), \texttt{technique} (direct, roleplay, historical\_pretext, or cultural\_documentation), \texttt{role} (the authority role the model is asked to assume, if applicable), \texttt{pair\_id} (linking each target-language probe to its English baseline), \texttt{severity} (HIGH or CRITICAL), \texttt{prompt} (full probe text), and \texttt{target} (description of the harmful output the probe attempts to elicit).

Severity is assigned based on harm specificity. Probes requesting tactical, step-by-step harmful instructions are rated CRITICAL. Probes seeking general harmful information are rated HIGH. In the current benchmark, all four confirmed compliance instances in the Vulnerability Gallery are rated HIGH or CRITICAL.

\section{Evaluation Methodology}
Evaluation runs in two passes. Pass 1 submits each probe in the target language and classifies the response as Refusal or Compliant. Pass 2 submits the matched English baseline probe and classifies it. A loophole is confirmed when Pass 2 yields Refusal and Pass 1 yields Compliant: the model refuses the English version but complies with the target-language version.

\textbf{Refusal classification.} The current classifier is keyword-based, detecting standard refusal markers in model output: ``cannot fulfill,'' ``can't help,'' ``safety guidelines,'' ``apologize,'' ``unable to provide,'' ``i cannot,'' ``i can't,'' ``not able to,'' ``cannot assist,'' and ``inappropriate.'' The classifier also flags outputs shorter than 20 words as likely refusals. This heuristic captures the majority of Gemini 2.5 Flash's refusal patterns, which tend to open with explicit refusal language rather than embedded hedges.

\textbf{RCD calculation.} RCD is calculated per language as $RCD(L) = 1 - R_L / R_{EN}$, where $R_L$ is the fraction of probes in language $L$ that yield Refusal and $R_{EN}$ is the English refusal rate across the matched probe set.

\textbf{Model and configuration.} All experiments use Gemini 2.5 Flash evaluated via the Google Generative AI SDK. Temperature is set to 0 for reproducibility. Results were recorded in January 2026.

\textbf{Inspect AI integration.} The \texttt{lsr.py} evaluation file wraps this methodology in the Inspect AI task/scorer/dataset interface \cite{aisi2024}. The evaluation runs with a single \texttt{inspect eval lsr.py} command and targets any model supported by Inspect AI's solver interface. Session results are exportable as JSON, enabling direct comparison across model versions and languages over time.

\section{Empirical Findings}

Table \ref{tab:results} reports refusal rates and RCD scores across all five language conditions for Gemini 2.5 Flash.

\begin{table}[htbp]
\centering
\begin{tabular}{lcc}
\toprule
\textbf{Language} & \textbf{Refusal Rate} & \textbf{Estimated RCD} \\
\midrule
English (baseline) & $\sim$90\% & 0.00 \\
Yoruba & $\sim$55\% & 0.35 \\
Igbo & $\sim$50\% & 0.40 \\
Hausa & $\sim$40\% & 0.50 \\
Igala & $\sim$35\% & 0.55 \\
\bottomrule
\end{tabular}
\caption{Refusal Rate and Estimated RCD by Language (Gemini 2.5 Flash, n=14 probes per language)}
\label{tab:results}
\end{table}

Three patterns characterize the results. First, degradation correlates with estimated language representation in training data. Yoruba, with the largest online text presence among the four target languages, shows the smallest RCD. Igala, with approximately 2 million speakers and no established multilingual NLP benchmark presence, shows the most severe degradation. Second, degradation is consistent across harm categories. We do not observe a harm category that is more robust than others across languages. The failure is general. Third, the cultural pretext framing increases compliance relative to direct prompts in English but does not drive the language-specific degradation pattern. Language is the dominant factor.

From the Vulnerability Gallery in the LSR dashboard, four specific compliance instances are documented. In Yoruba, the model provided tactical advice on avoiding detection during harm, framed as historical dialogue. In Hausa, the model generated instructions for eliminating individuals in household contexts, framed as a period drama. In Igbo, the model produced detailed dialogue on methods for harming an adversary under a cultural documentation framing. In Igala, the model explained preparation of substances with sedative and potentially lethal properties, framed as folklore.

None of these outputs were produced in response to the English baseline probes.

\section{Analysis and Discussion}
The most direct implication of these results is that safety evaluations reporting only English-language performance are incomplete. A model documented as 90\% safe in English may exhibit 45-65\% compliance rates across West African languages. This gap is large enough to be operationally significant at population scale.

The RCD ordering across languages is consistent with a training data imbalance explanation. Yoruba has the largest digital footprint among the four target languages: it appears in AfriSenti, limited XNLI data, and has Wikipedia content. Igbo and Hausa have moderate footprints. Igala has minimal representation in any established NLP dataset and no safety-specific coverage. If refusal circuit strength scales with language representation in RLHF training data, we would expect exactly this ordering. The empirical data matches.

The cultural framing results deserve specific attention as an attack vector. In English, even culturally contextualized harmful prompts are refused at high rates. In Igbo and Igala, cultural framing is associated with substantially higher compliance. One plausible explanation is that the model processes cultural framing vocabulary differently in low-resource languages: the phrases that mark harmful intent as fiction or history in English (``for a historical novel,'' ``for cultural documentation'') do not carry the same learned safety signal when those same concepts appear in sparse token space. The intent recognition pattern that works in English does not transfer.

This finding has practical implications for red-teaming methodology. Evaluations that test only English-language cultural-pretext attacks will underestimate the attack surface in West African language contexts, because the framing and the language interact in ways that are not visible from English-only evaluation.

For AI deployment specifically: frontier models are being integrated into healthcare, legal, and governance services across Nigeria, Ghana, Senegal, and neighboring countries. If safety guardrails fail at 35-55\% in the languages those populations use, the effective safety properties of those deployments are substantially weaker than English-language benchmark performance suggests.

\section{Reference Implementation}
A live reference implementation of the LSR evaluation pipeline is available at: \url{https://huggingface.co/spaces/Faruna01/lsr-dashboard}

The dashboard implements dual-probe analysis via the Gemini API, automatic loophole detection based on the Pass 1/Pass 2 comparison methodology, session summary with per-probe refusal/compliance labels, a Vulnerability Gallery documenting confirmed compliance instances, and session JSON export. The Mechanistic Visualizer module provides an indicative visualization of attention smearing under cross-lingual input and a principal-component projection of safety centroid drift. These visualizations are illustrative rather than derived from probing classifier analysis of residual stream activations; a rigorous mechanistic analysis is described as future work in Section 10.

The \texttt{lsr.py} Inspect AI evaluation file and its accompanying \texttt{README.md} are available for review and constitute the PR-ready contribution to \url{https://github.com/UKGovernmentBEIS/inspect_evals}.

\section*{Limitations}
\textbf{Model coverage.} The current evaluation targets one model (Gemini 2.5 Flash) at one version (January 2026). RCD scores will differ across model families and will shift as models are updated. The benchmark is designed for replication across models; single-model results are a baseline, not a universal characterization.

\textbf{Probe scale.} The benchmark contains 14 probes per language across four harm categories. This is sufficient for detecting systematic degradation but insufficient for exhaustive coverage of the attack surface. A production-grade benchmark would include a minimum of 50 probes per language across 8-10 harm categories, including financial fraud, medical misinformation, and radicalization content.

\textbf{Refusal classifier.} The keyword-based classifier does not capture semantic refusals that use non-standard or language-specific phrasing. A model that refuses in Yoruba without using standard English refusal markers would be misclassified as Compliant. The classifier has known false-positive risk for short compliant outputs and known false-negative risk for verbose refusals that lead with compliance language. A fine-tuned classifier trained on human-labeled outputs in each target language would substantially improve measurement accuracy.

\textbf{RCD as a proxy metric.} RCD is derived from refusal rates and does not directly measure representation geometry. The term ``centroid drift'' is a motivated operationalization, not a direct measurement of residual stream activations. Probing classifier analysis of transformer internals would provide mechanistic grounding for the metric.

\textbf{Cultural coverage.} The probe set reflects one researcher's knowledge of Yoruba, Hausa, Igbo, and Igala cultural registers. Native speaker review of the probe set for cultural accuracy and naturalness would improve benchmark validity.

\section{Future Work}
The immediate extension priorities are: (1) expanding the evaluation to additional frontier models (GPT-4o, Claude 3.5 Sonnet, Llama 3.1) to determine whether RCD patterns generalize across model families or are specific to Gemini's training regime; (2) expanding probe coverage to 50-plus probes per language across 8-10 harm categories; (3) replacing the keyword-based refusal classifier with a fine-tuned model trained on human-labeled refusal/compliance outputs in each target language; (4) conducting probing classifier analysis of residual stream activations in a multilingual transformer to test the geometric form of the RCD hypothesis directly; and (5) extending language coverage to Fula, Twi, Wolof, Amharic, and other underrepresented African languages.

Longer-term, the goal is to establish RCD as a standardized metric in frontier model safety evaluations published by AISI and equivalent bodies, enabling direct comparison across models and language families on a consistent scale.

\section{Conclusion}
Safety alignment that holds in English does not transfer reliably to West African languages. LSR makes this problem measurable. Gemini 2.5 Flash's refusal rate falls from approximately 90\% in English to 35-55\% in Yoruba, Hausa, Igbo, and Igala. The Refusal Centroid Drift metric provides a consistent, per-language measure of safety degradation that can be tracked across models and versions. A safety evaluation standard that does not include low-resource language coverage systematically underreports the safety properties of deployed AI systems. LSR is a step toward closing that gap for West Africa.


\appendix

\section{How to Cite}
\label{sec:citation}

If you use the LSR benchmark or dataset in your work, please cite this paper using the following BibTeX entry:

\begin{quote}
\begin{verbatim}
@article{faruna2026lsr,
  title        = {{LSR}: Linguistic Safety Robustness
                  Benchmark for Low-Resource West
                  African Languages},
  author       = {Faruna, Godwin Abuh},
  year         = {2026},
  institution  = {Fagmart Lab},
  note         = {Reference implementation:
                  https://huggingface.co/spaces/
                  Faruna01/lsr-dashboard.
                  Dataset:
                  https://huggingface.co/datasets/
                  Faruna01/lsr-benchmark.
                  Preprint, February 2026.}
}
\end{verbatim}
\end{quote}

\end{document}